\documentclass[letterpaper, 10 pt, conference]
{ieeeconf}  

\IEEEoverridecommandlockouts                              

\usepackage{xcolor}
\usepackage{amsmath,amssymb,amsfonts,color,nicefrac}
\usepackage{bbm}
\usepackage{graphicx}
\usepackage{algorithm}
\usepackage{algpseudocode}
\usepackage{comment}
\usepackage{caption}
\usepackage{subcaption}
\usepackage{cite}
\DeclareMathOperator*{\argmin}{arg\,min}

\DeclareSymbolFont{matha}{OML}{txmi}{m}{it}
\DeclareMathSymbol{\varv}{\mathord}{matha}{118}
\renewcommand{\Re}{\mathbb{R}}

\newtheorem{proposition}{Proposition}
\newtheorem{theorem}{Theorem}

\title{\LARGE \bf
Distributed Event-Triggered Distance-Based \\ Formation Control for Multi-Agent Systems 
}

\author{Evangelos Psomiadis and Panagiotis Tsiotras
\thanks{The work was supported by the ARL grant DCIST CRA W911NF-17-2-0181.}
\thanks{E. Psomiadis and P. Tsiotras are with the D. Guggenheim School of Aerospace Engineering, Georgia Institute of Technology, Atlanta, GA, 30332-0150, USA. Email:
       \{epsomiadis3,tsiotras\}@gatech.edu}%
}

\begin{document}
\maketitle
\thispagestyle{empty}
\pagestyle{empty}

\begin{abstract}
This paper addresses collaborative formation control for multi-agent systems with limited resources.
We consider a team of robots tasked with achieving a desired formation from an arbitrary initial configuration. 
To reduce unnecessary control updates and conserve resources, we propose a distributed event-triggered formation controller.
Unlike the well-studied linear formation controllers, the proposed controller is nonlinear and relies on inter-agent distance measurements.
Control updates are triggered only when the measurement error exceeds a state-dependent threshold, ensuring system stability while reducing actuation effort. 
We also employ a distributed control barrier function to guarantee inter-agent collision avoidance.
The proposed controller is validated through simulations and real-world experiments involving different communication topologies, scalability tests, and variations in design parameters, and is also compared to periodic triggering strategies.
Results demonstrate that the event-triggered approach significantly reduces control effort while preserving formation performance.
\end{abstract}

\section{Introduction}

Multi-agent robotic systems often collaborate autonomously by leveraging individual capabilities and inter-agent interactions to achieve system-level objectives.
These objectives often involve consensus, flocking, and formation control~\cite{Fax2004_formation}. 
Networked control theory has emerged as a powerful tool for enabling such tasks \cite{Mesbahi2010}.
Recent research has focused on refining these controllers by accounting for the robots' inherent resource limitations.

As robotic systems grow more complex, reducing unnecessary actuator updates is crucial for conserving energy and easing the load on embedded processors and communication buses.
A viable approach to achieve this objective is \textit{event-triggered control}, where control tasks are scheduled based on events rather than conventional periodic sampling.
It has been shown \cite{JOHANASTROM1999} that, under certain conditions, event-triggered controllers can outperform time-triggered ones.
In \cite{Tabuada2007}, the authors introduced an event-triggered strategy dependent on the state norm.
Since then, many researchers have further investigated this topic by proposing optimal event-triggered schemes \cite{Maity2020} or data-driven approaches \cite{DePersis_2024}.

Beyond single-agent systems, event-triggered controllers have also been developed for networked control systems \cite{NOWZARI20191}, with several applications, including UAV light shows \cite{Zhang2023} and connectivity preservation for unmanned surface vehicles \cite{Chen2024}. 
In \cite{Dimarogonas2012_ET_MA}, the authors introduced event-triggered consensus controllers.
In \cite{SEYBOTH2013245}, the authors proposed an event-based broadcasting strategy.
A direct extension of consensus is displacement-based formation control.
The authors in \cite{Li2017, Toyota2018} proposed event-triggered formation controllers over displacements for continuous and discrete dynamics, respectively, while \cite{Yi2017} considered connectivity preservation.
In \cite{Chen2024} and \cite{Guo2025}, this idea was extended to leader–follower settings.

Nevertheless, displacement-based controllers are unsuitable in many scenarios because they require agents to share a consistent orientation or a common reference frame \cite{Li2017, Toyota2018}.
Moreover, in many cases, the control objective is not to realize a specific geometric formation but rather to maintain prescribed distances for purposes such as ensuring safety or preserving network connectivity.
To address such tasks, distance-based formation controllers have been developed \cite{Krick2008, SAHEBSARA2024}. 
Although they are more difficult to analyze than simple consensus or displacement-based controllers due to their inherent nonlinearity, they offer significant advantages. 
In particular, distance-based controllers eliminate the need for a global reference frame, allowing the desired formation to be realized through multiple equivalent translations and rotations.
For example, \cite{Liu2015} proposed a centralized event-triggered distance-based formation controller, while \cite{Yu2012_ev-tr-formation} presented a distributed distance-based event-triggered strategy.
It is worth noting, however, that in \cite{Yu2012_ev-tr-formation} the event-triggering mechanism governs only the transmission of state information, not the update times of the control inputs.

\begin{figure}[tb]
     \centering
     \begin{subfigure}[b]{0.48\linewidth}
         \centering
         \includegraphics[width=1\linewidth]{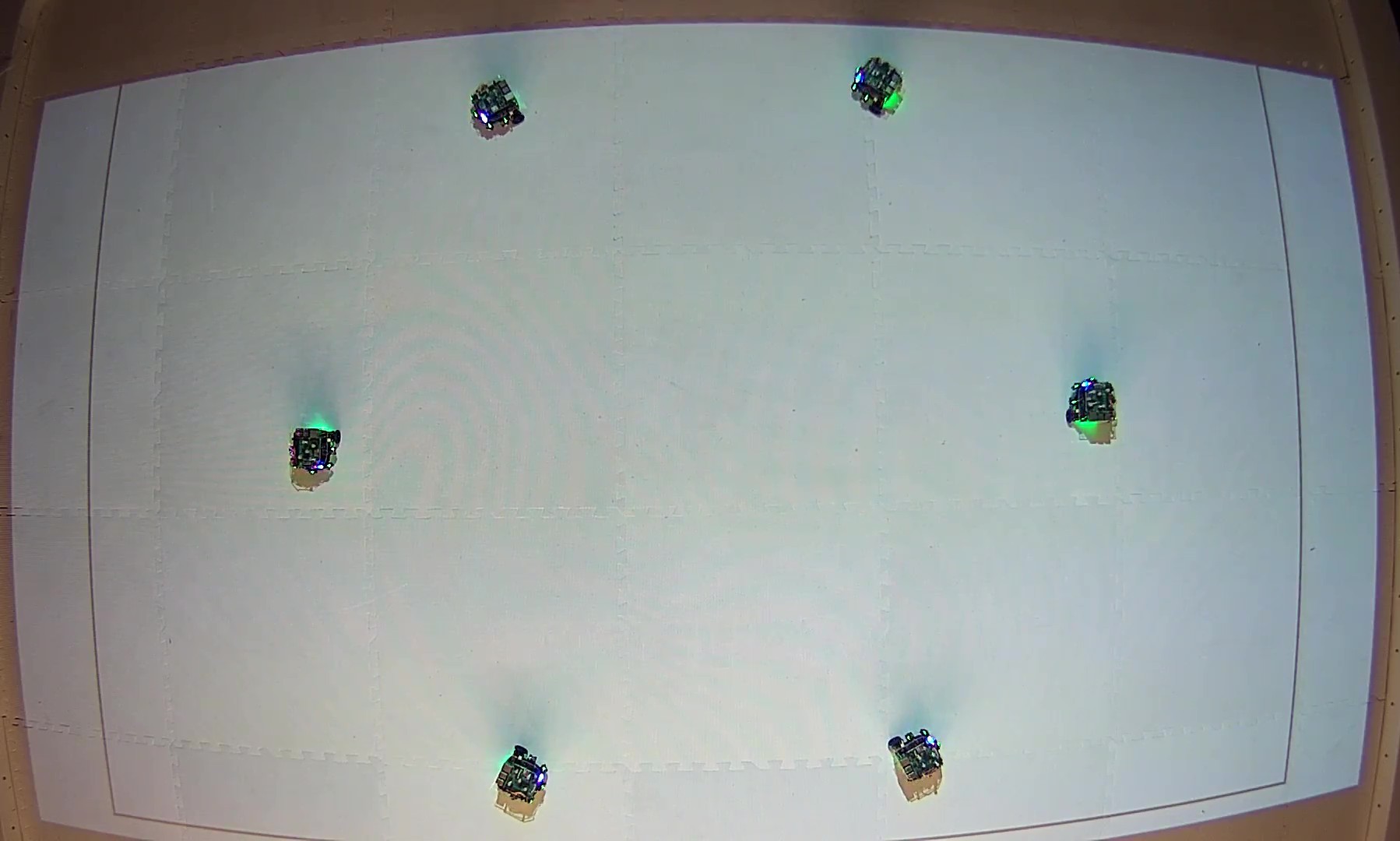}
         \caption{}
         \label{fig:robotarium_in}
     \end{subfigure}
     \begin{subfigure}[b]{0.48\linewidth}
         \centering
         \includegraphics[width=1\linewidth]{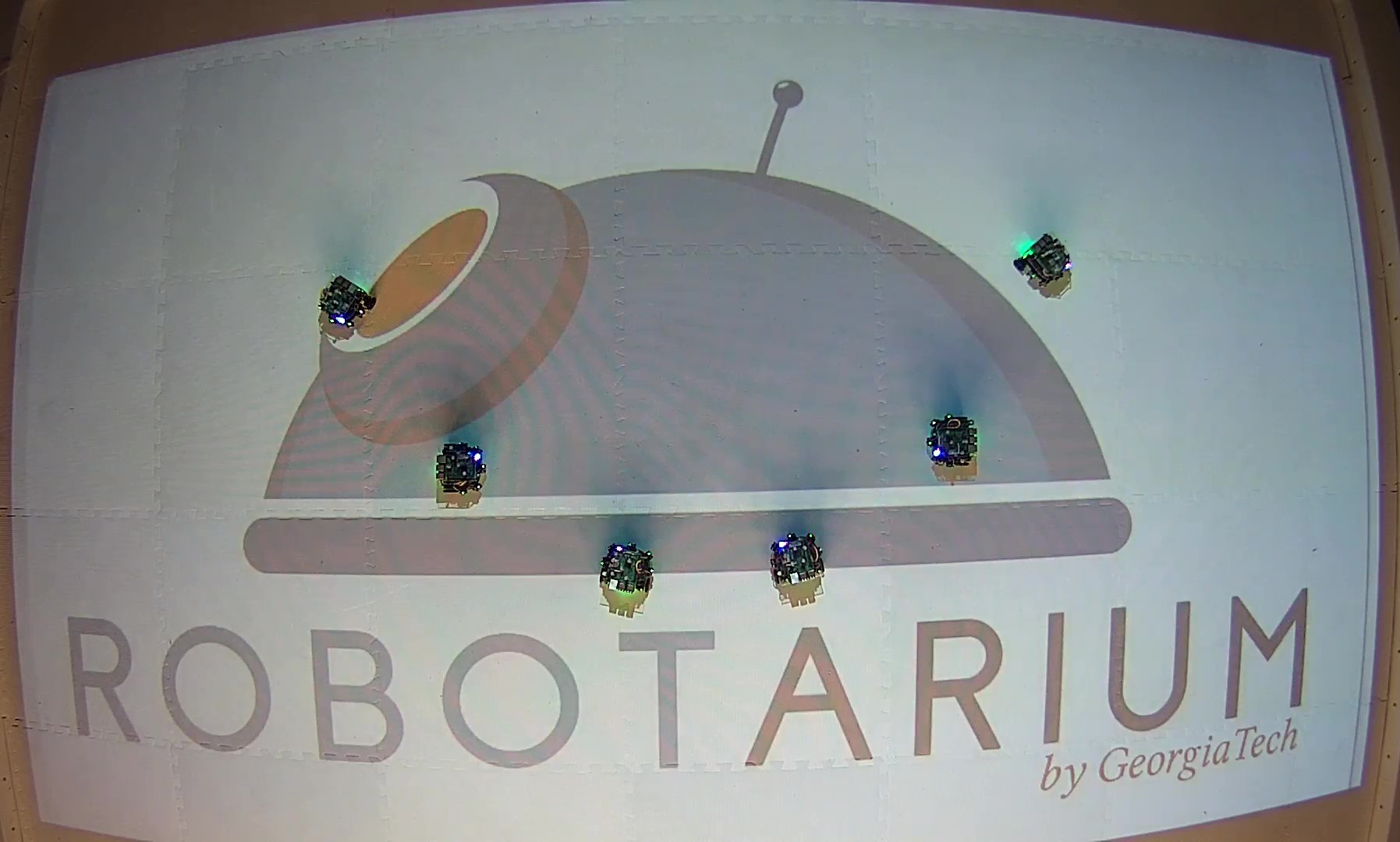}
         \caption{}
         \label{fig:robotarium_fin}
     \end{subfigure}
        \caption{Six GRITSBot X robots, initially arranged in a circular formation (a), achieve a V-formation (b) by employing the proposed event-triggered controller in the Robotarium \cite{Wilson2020_robotarium}.} 
        \label{fig:robotarium}
\end{figure}

To the best of our knowledge, distributed distance-based event-triggered formation control schemes that determine when the control inputs should be updated remain unexplored.
This is the main contribution of this paper.
Moreover, we explicitly incorporate collision avoidance through a distributed control barrier function framework~\cite{Ames2019,Wang2017}. Finally, we conduct several experiments and simulations using the Georgia Tech Robotarium simulator and hardware testbed \cite{Wilson2020_robotarium}, as well as large-scale simulations of 200 agents.

\section{Notation and Preliminaries} \label{sec:preliminaries}

Lowercase and uppercase bold symbols denote vectors and matrices, respectively. 
$\|\! \cdot \!\|$ denotes the Euclidean norm.
Let $\mathbf{x}$ be the stacked vector representing a variable across all agents.
The subscript $i$ in $\mathbf{x}_{i}$ indicates the component associated with agent $i$.
Unless otherwise specified, $\mathbf{x}_{i}$ depends on time $t$, but we suppress the explicit argument notation for conciseness.
We use the subscript $ij$ in $\mathbf{A}_{ij}$ to denote the ($i,j$)-th element of $\mathbf{A}$.
Let $G = (V, E)$ be an undirected graph, where $V = \{1, \dots, N\}$ is the set of nodes, and $E$ is the set of edges.  
$G$ is called (infinitesimally) rigid in $\Re^\mathcal{D}$ if, for a given placement of its nodes in space, the only infinitesimal motions that preserve the distances between connected vertex pairs are trivial rigid-body motions (translations and rotations) \cite{Mesbahi2010}.
For a formal analysis and definition of graph rigidity, the interested reader is referred to \cite{ASIMOW1979}.
The set of neighbors of node $i$ is denoted as $\mathcal{N}_i = \{ j \in V \mid (i,j) \in E \}$,  
and the degree of node $i$ is defined as $|\mathcal{N}_i|$, the cardinality of $\mathcal{N}_i$.
The adjacency matrix $\mathbf{A} \in \mathbb{R}^{N \times N}$ of the graph $G$ is defined such that $\mathbf{A}_{ij} = 1$, if $(i,j) \in E$, and $\mathbf{A}_{ij} = 0$ otherwise.
The degree matrix $\mathbf{R}  \in \mathbb{R}^{N \times N}$ is diagonal, with each diagonal element equal to the degree of the corresponding node.
The Laplacian matrix of $G$, defined as $\mathbf{L} = \mathbf{R}  - \mathbf{A}$, describes the network connectivity and is symmetric and positive semidefinite.
Throughout this work, we assume that all graphs under consideration are connected.
We also define the corresponding incidence matrix $\mathbf{B} \in \Re^{N \times M}$ of the Laplacian by $\mathbf{L} =\mathbf{B}\mathbf{B}^\top$, where $M = |E|$.

\section{Problem Formulation}

\begin{figure}[tb]
    \centering        
    {\includegraphics[width=0.85\linewidth]{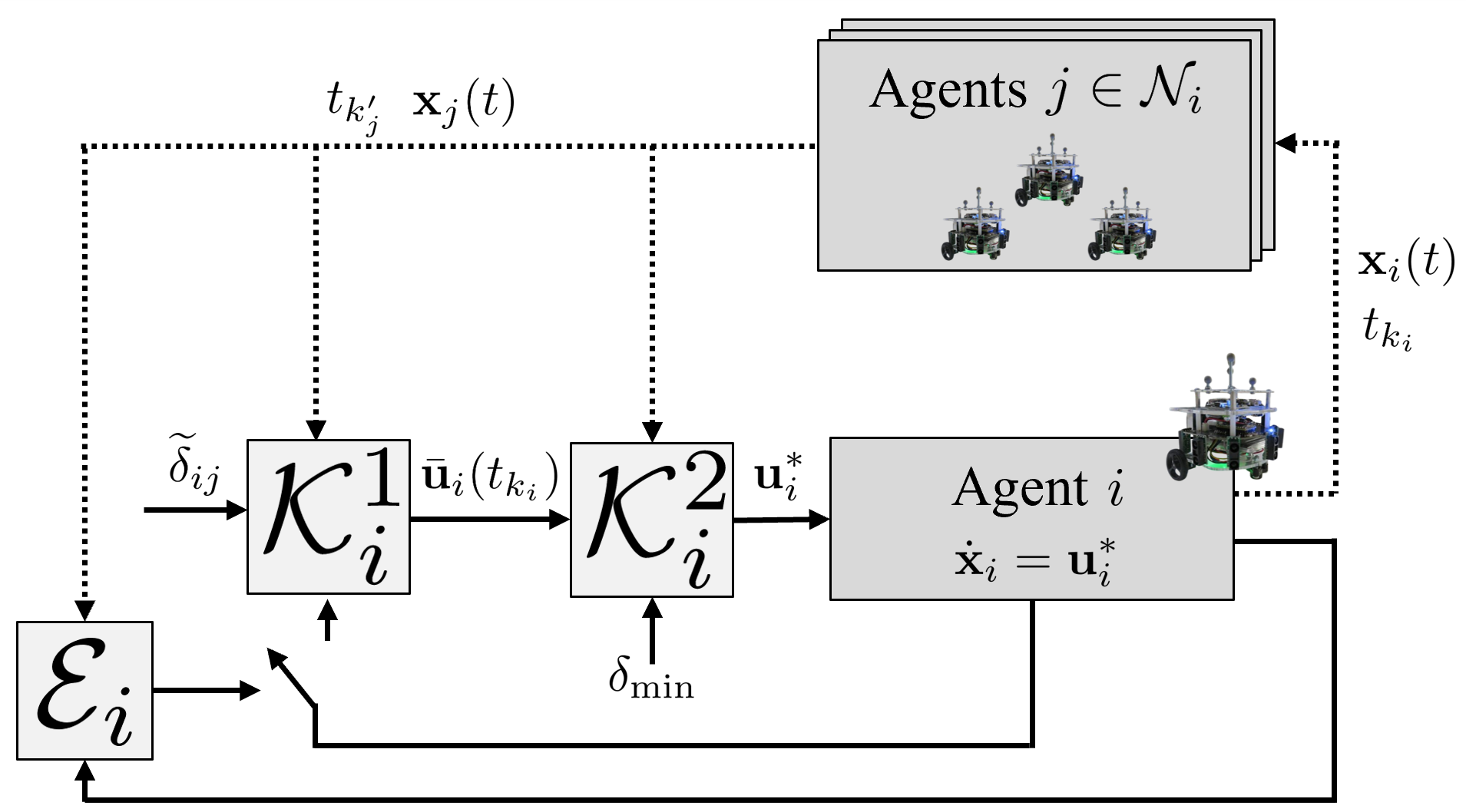}}
    \caption{Distributed, event-triggered, distance-based formation control loop of agent $i$.} 
    \label{fig:framework}
\end{figure}

Consider a team of $N \geq 2$ agents moving in $\Re^\mathcal{D}$, $\mathcal{D} \in\{2,3\}$, where the position of agent $i$ is denoted by $\mathbf{x}_i \in \Re^\mathcal{D}$ and $\mathbf{x} \in \Re^{\mathcal{D}N}$ is the stacked vector of agents' positions. 
Assume that no two agents are collocated initially.
The dynamics of each agent are
\begin{equation} \label{eq:dynamics}
    \dot{\mathbf{x}}_i = \mathbf{u}_i, \quad i = 1, \dots, N,
\end{equation}
where $\mathbf{u}_i$ denotes the control input for agent $i$.

The communication network of the agents is modeled as an undirected graph $G = (V,E)$, where the set of nodes $V = \{1, \ldots, N\}$ corresponds to the agents, and the set of edges $E$ represents communication links.

The system-level objective is for the agents to achieve a desired translation- and rotation-invariant formation while reducing the frequency of the control input updates through event-triggering.
Let $\delta_{ij}$
denote the relative distance between agents $i$ and $j$, defined as $\delta_{ij}(\mathbf{x}) = \| \mathbf{x}_j - \mathbf{x}_i\|$.
We define a formation based on distances as the set  
\begin{equation} \label{eq:form_distances}
    X = \{\mathbf{x} \in \mathbb{R}^{\mathcal{D}N} : \|\mathbf{x}_j - \mathbf{x}_i\| = \widetilde{\delta}_{ij}, \quad \forall (i,j) \in E\},
\end{equation}
where $\widetilde{\delta}_{ij}$ is the desired distance between agents $i$ and $j$.
Note that the desired distances $\widetilde{\delta}_{ij}$ must define a feasible formation and satisfy $\widetilde{\delta}_{ij}>\delta_{\min}$, so that the formation control and collision avoidance objectives do not conflict.

To reduce actuator updates, we adopt an event-triggered controller that selects the update times based on a triggering condition. 
We propose a distributed, event-triggered, distance-based formation controller with inter-agent collision avoidance.
A schematic of the proposed controller is presented in Figure~\ref{fig:framework}, where $\mathcal{E}_i$ is the event generator of agent $i$, $\mathcal{K}_i^1$ represents the event-triggered (nominal) controller, and $\mathcal{K}_i^2$ is responsible for collision avoidance.
Note that agent $i$ can measure its neighbor's state $\mathbf{x}_j$ using a camera or LiDAR, and that agent $j$ can signal its triggering time.

\section{Distributed Event-triggered Formation Control}

In this section, we introduce the proposed event-triggered controller.  
Let $t_{k_i}$ denote the $k$-th trigger's execution time associated with the update of the control input of agent $i$.
The measurement error is defined as $\mathbf{e}_i (t) = \mathbf{x}_i(t_{k_i}) - \mathbf{x}_i(t), ~ i = 1, \dots, N$ for $t \in [t_{k_i},t_{k_i + 1})$.
In our framework, each agent $i$ measures its neighbors' positions and receives their execution times $t_{k_j}$ for all $j \in \mathcal{N}_i$.
Using this information, agent $i$ can compute its neighbors' measurement errors.

Each agent updates its nominal control input $\bar{\mathbf{u}}_i$ only at discrete time instants $t_{k_i}$.
Between control updates, the control input $\bar{\mathbf{u}}_i$ remains constant, i.e., $\bar{\mathbf{u}}_i(t) = \bar{\mathbf{u}}_i(t_{k_i})$ for all $t \in [t_{k_i},t_{k_i + 1}).$
The nominal control input for agent $i$ is updated at $t_{k_i}$ as follows
\begin{align} \label{eq:distributed_ctrl_i}
\bar{\mathbf{u}}_i(t_{k_i}) = - \alpha\sum_{j \in \mathcal{N}_i} ( {\delta}_{ij}^2 - \widetilde{\delta}_{ij}^2) (\mathbf{x}_i(t_{k_i}) - \mathbf{x}_j(t_{k_j})),
\end{align}
where $\alpha >0$ is a constant that depends on the maximum velocity bounds, 
${\delta}_{ij} (\mathbf{x}(\mathbf{t_k})) = \| \mathbf{x}_j(t_{k_j}) - \mathbf{x}_i(t_{k_i})\|$, 
$k_j = \argmin_{l_j \in \mathbb{N}, t \geq t_{l_j}} (t - t_{l_j})$ 
corresponds to the latest update of agent $j$, and, with a slight abuse of notation, $\mathbf{t_k}$ denotes the stacked vector of $t_{k_j}$ for $j = 1, \dots, N$, that is, $\mathbf{t_k}$ consists of the most recent control-update times of all agents.

We can write the controller in a more compact form by defining the matrix $\mathbf{W}(\mathbf{x}) \in \mathbb{R}^{M \times M}$, 
containing all inter-agent distances, as $\mathbf{W}(\mathbf{x}) = \mathrm{diag} ( \{ \delta_{\epsilon}^2(\mathbf{x}) \}_{\epsilon \in E} )$ and a similar diagonal matrix $\widetilde{\mathbf{W}}$ constructed using the desired edge distances $\widetilde{\delta}_{\epsilon}$. 
Then, the stacked vector of the control is
\begin{align} \label{eq:distributed_ctrl}
\bar{\mathbf{u}}(\mathbf{t_k}) 
&= - \alpha \, \widehat{\mathbf{L}}\big( \mathbf{x}(\mathbf{t_k}) \big) \, \mathbf{x}(\mathbf{t_k}),
\end{align}
where the weighted Laplacian matrix is $\widehat{\mathbf{L}}(\mathbf{x}) = \mathbf{L}(\mathbf{x}) \otimes I_{\mathcal{D}}$ with $\mathbf{L}(\mathbf{x}) = \mathbf{B} \big( \mathbf{W}(\mathbf{x}) - \widetilde{\mathbf{W}} \big) \mathbf{B}^\top.$

Hence, using \eqref{eq:dynamics}, the closed-loop system is
\begin{align} \label{cl_sys}
    \dot{\mathbf{x}}(t) = - \alpha \widehat{\mathbf{L}}(\mathbf{x}(t) + \mathbf{e}(t)) (\mathbf{x}(t) + \mathbf{e}(t)),
\end{align}
where $\mathbf{e}$ is the stacked vector of measurement errors.

\subsection{Stability}

The following theorem establishes the stability of the closed-loop system under the proposed nominal controller.

\begin{theorem}\label{thm:distributed_ctrl}
Consider the system $\dot{\mathbf{x}} = \bar{\mathbf{u}}$ under the control law \eqref{eq:distributed_ctrl}, where the communication graph $G$ is connected and rigid and the desired inter-agent distances encoded in $\widetilde{\mathbf{W}}$ correspond to a formation as in \eqref{eq:form_distances}. 
Suppose that each agent $i$ updates its control input whenever the following condition holds for at least one neighbor $j\in\mathcal{N}_i$:
\begin{align} \label{event_distr}
 \| \mathbf{e}_{j} - \mathbf{e}_{i}\| \geq (\sqrt{\beta_i (\mathbf{x}) + 1} - 1) {\delta}_{ij}(\mathbf{x}) ,
\end{align}
where,
\begin{align} \label{eq:beta_i}
    \beta_i(\mathbf{x}) = \sigma_i \left( \sqrt{ \frac{\|\mathbf{z}_i(\mathbf{x})\|}{\sum_{j \in \mathcal{N}_i}\delta_{ij}^3(\mathbf{x})} +  B_i^2 (\mathbf{x})} - B_i 
    (\mathbf{x}) \right),
\end{align}
\begin{align} \label{eq:B_i}
    B_i(\mathbf{x} ) = \frac{1}{2} \left(\frac{ \sum_{j \in \mathcal{N}_i}\delta_{ij}(\mathbf{x}) |D_{ij}(\mathbf{x})| }{\sum_{j \in \mathcal{N}_i}\delta_{ij}^3(\mathbf{x})} + 1\right),
\end{align}
$\mathbf{z}_i(\mathbf{x}) = \widehat{\mathbf{L}}_i(\mathbf{x}) \mathbf{x} = \sum_{j \in \mathcal{N}_i} D_{ij}  (\mathbf{x}_i - \mathbf{x}_j)$, 
$D_{ij}(\mathbf{x} ) = {{\delta}_{ij}^2(\mathbf{x} ) - \widetilde{\delta}_{ij}^2}$, and
$0 < \sigma_i < 1$.
Then, the inter-agent distance errors locally asymptotically converge to zero, i.e.,
\begin{equation}
    \lim_{t\to\infty} D_{ij} = 0,
    \qquad \forall (i,j)\in\mathcal{E}.
\end{equation}
\end{theorem}

\begin{proof}
Consider the candidate Lyapunov function
\begin{equation} \label{eq:LF}
    V(\mathbf{x}) = \frac{1}{8\alpha } \sum_{i = 1}^N \sum_{j \in \mathcal{N}_i} D_{ij}^2,
\end{equation}
where $D_{ij}(\mathbf{x}) = \delta_{ij}^2(\mathbf{x}) - \widetilde{\delta}_{ij}^2$.
Taking the gradient of $V$ yields
$\frac{\partial V} {\partial \mathbf{x}_i} = \frac{1}{\alpha} \sum_{j \in \mathcal{N}_i} D_{ij}  (\mathbf{x}_i - \mathbf{x}_j) = \frac{1}{\alpha} \widehat{\mathbf{L}}_i(\mathbf{x}) \mathbf{x}.$
Note that the elements of the Laplacian are given by
\begin{equation} \label{eq:Laplacian_elements}
\mathbf{L}_{ij}(\mathbf{x}) =
\begin{cases}
\sum_{q \in \mathcal{N}_i} D_{iq}, & \text{if } i=j, \\
- D_{ij}, & j \in \mathcal{N}_i, \\
0, & \text{otherwise}.
\end{cases}
\end{equation}

Hence, we can write 
$ \widehat{\mathbf{L}}(\mathbf{x} + \mathbf{e}) = \widehat{\mathbf{L}}(\mathbf{x}) + \widehat{\mathbf{C}}(\mathbf{x}, \mathbf{e}),$
where $\widehat{\mathbf{C}} = \mathbf{C} \otimes I_{\mathcal{D}}$ and $\mathbf{C}$ captures the variation due to the measurement error. 
Its off-diagonal elements are the negative of $E_{ij}(\mathbf{x}, \mathbf{e}) = \|\mathbf{e}_j - \mathbf{e}_i\|^2 + 2 (\mathbf{x}_j - \mathbf{x}_i)^\top (\mathbf{e}_j - \mathbf{e}_i) $ if $j \in \mathcal{N}_i$, and $0$ otherwise; the diagonal entries are defined to ensure zero row sum.
By taking the time derivative of \eqref{eq:LF}, we have
\begin{align} \label{eq:LF_distr_deriv2}
    \dot{V} &= \frac{\partial V} {\partial \mathbf{x}} \dot{\mathbf{x}} = - \mathbf{x}^\top \widehat{\mathbf{L}}(\mathbf{x})  \widehat{\mathbf{L}}(\mathbf{x} + \mathbf{e}) (\mathbf{x} + \mathbf{e} ) 
    \nonumber \\
    &= - \mathbf{x}^\top \widehat{\mathbf{L}}(\mathbf{x})  \widehat{\mathbf{L}}(\mathbf{x}) (\mathbf{x} + \mathbf{e} )  -  \mathbf{x}^\top \widehat{\mathbf{L}}(\mathbf{x})  \widehat{\mathbf{C}}(\mathbf{x}, \mathbf{e}) \mathbf{x}
    \nonumber \\
    & \quad -  \mathbf{x}^\top \widehat{\mathbf{L}}(\mathbf{x})  \widehat{\mathbf{C}}(\mathbf{x}, \mathbf{e}) \mathbf{e}.
\end{align}
Let $\mathbf{z} = \widehat{\mathbf{L}}(\mathbf{x}) \mathbf{x}$.
Using \eqref{eq:Laplacian_elements}, the first term of the right-hand side (RHS) of \eqref{eq:LF_distr_deriv2} can be written as 
\begin{align} \label{eq:RHS1}
-\sum_i \mathbf{z}_{i}^2 -\sum_i \sum_{j \in \mathcal{N}_i} D_{ij}   ( \mathbf{e}_{i} - \mathbf{e}_{j} )^\top  \mathbf{z}_{i}.
\end{align}
From the event-triggering condition in \eqref{event_distr},
while no event is triggered for agent $i$, it follows that
\begin{align} \label{event_1_distr_ineq}
     \| \mathbf{e}_{j} - \mathbf{e}_{i}\| \leq (\sqrt{\beta_i + 1} - 1) \| \mathbf{x}_{j} - \mathbf{x}_{i} \|, \quad \forall j \in \mathcal{N}_i,
\end{align}
where $\beta_i>0$ is a scalar.
From the triangle and the Cauchy–Schwarz inequalities, note that $| E_{ij}| = | \|\mathbf{e}_j - \mathbf{e}_i\|^2 + 2 (\mathbf{x}_j - \mathbf{x}_i)^\top (\mathbf{e}_j - \mathbf{e}_i) | \leq \beta_i\|\mathbf{x}_j - \mathbf{x}_i\|^2 = \beta_i {\delta}_{ij}^2$.
Using \eqref{event_1_distr_ineq}, the expression in \eqref{eq:RHS1} can be bounded from above by 
\begin{align*}
     -\sum_i \mathbf{z}_i^2 + \sum_i (\sqrt{\beta_i + 1} - 1) \sum_{j \in \mathcal{N}_i} \delta_{ij} |D_{ij}|  \|\mathbf{z}_i\|.
\end{align*}
The second term of the RHS of \eqref{eq:LF_distr_deriv2} can be written as
\begin{align*}
    - \sum_i\sum_{j \in \mathcal{N}_i} E_{ij} (\mathbf{x}_i - \mathbf{x}_j)^\top \mathbf{z}_i \leq
     \sum_i \beta_i \sum_{j \in \mathcal{N}_i} \delta_{ij}^3 \|\mathbf{z}_i\|.
\end{align*}
The third term of the RHS of \eqref{eq:LF_distr_deriv2} can be upper-bounded by 
\begin{align*}
    &-  \mathbf{x}^\top \widehat{\mathbf{L}}(\mathbf{x})  \widehat{\mathbf{C}}(\mathbf{x}, \mathbf{e}) \mathbf{e} \leq
        \sum_i \beta_i (\sqrt{\beta_i + 1} - 1) \sum_{j \in \mathcal{N}_i} \delta_{ij}^3 \| \mathbf{z}_i \|.
\end{align*}
Finally, it follows that
\begin{align} \label{eq:LF_distr_deriv3}
    \dot{V} & \leq   -\sum_i  A_{i}(\mathbf{x}, \mathbf{z}) \| \mathbf{z}_i \|, 
\end{align}
where 
$A_{i}(\mathbf{x}, \mathbf{z}) =  \|\mathbf{z}_i\| - (\sqrt{\beta_i + 1} - 1) \sum_{j \in \mathcal{N}_i} \delta_{ij} |D_{ij}| - \beta_i \sum_{j \in \mathcal{N}_i} \delta_{ij}^3 - \beta_i (\sqrt{\beta_i + 1} - 1) \sum_{j \in \mathcal{N}_i} \delta_{ij}^3$ or, 
\begin{align} \label{eq:A_i}
    A_{i}(\mathbf{x}, \mathbf{z}) &=  \|\mathbf{z}_i\| + \sum_{j \in \mathcal{N}_i} \delta_{ij} |D_{ij}| \nonumber \\
    &- \sqrt{\beta_i + 1} \left( \sum_{j \in \mathcal{N}_i} \delta_{ij} |D_{ij}|  + \beta_i \sum_{j \in \mathcal{N}_i} \delta_{ij}^3 \right).
\end{align}

From the condition of $\beta_i$ in~\eqref{eq:beta_i} and since
$\sigma_i<1$, we have
\begin{align} \label{eq:beta_proof}
    \beta_i < \sqrt{ \frac{\|\mathbf{z}_i\|}{\sum_{j \in \mathcal{N}_i}\delta_{ij}^3} +  B_i^2 } - B_i,
\end{align}
where $B_i = \frac{1}{2} \left(\frac{ \sum_{j \in \mathcal{N}_i}\delta_{ij} |D_{ij}| }{\sum_{j \in \mathcal{N}_i}\delta_{ij}^3} + 1\right)$.
Rearranging~\eqref{eq:beta_proof} and squaring both sides yields $(\beta_i + 1) ( \sum_{j \in \mathcal{N}_i} \delta_{ij} |D_{ij}|  + \beta_i \sum_{j \in \mathcal{N}_i} \delta_{ij}^3 ) < \|\mathbf{z}_i\| + \sum_{j \in \mathcal{N}_i} \delta_{ij} |D_{ij}|$.
Since $\sqrt{\beta_i+1}<\beta_i+1$,  we have $\sqrt{\beta_i + 1} ( \sum_{j \in \mathcal{N}_i} \delta_{ij} |D_{ij}|  + \beta_i \sum_{j \in \mathcal{N}_i} \delta_{ij}^3 ) < \|\mathbf{z}_i\| + \sum_{j \in \mathcal{N}_i} \delta_{ij} |D_{ij}|$. 
It then follows directly from~\eqref{eq:A_i} that $A_{i}(\mathbf{x}, \mathbf{z}) > 0$.
Moreover, $A_i \rightarrow 0$ if and only if $\|\mathbf{z}_i\| \rightarrow 0$, since in this case $\beta_i \rightarrow 0$, otherwise $\beta_i$ remains strictly positive.
Additionally, note that $A_i > 0$ when $\delta_{ij} = 0$.
However,  $\delta_{ij} = 0$ would imply that the robots are in consensus, which contradicts the objective of distance-based formation control, and it is forbidden due to the collision avoidance module (see Section \ref{sec:collision avoidance}). 
Hence, \eqref{eq:LF_distr_deriv3} results in the expression $\dot{V} \leq -\sum_i  A_{i} \| \mathbf{z}_i \| =    -\sum_i  A_{i} \| \widehat{\mathbf{L}}_i(\mathbf{x}) \mathbf{x} \|  \leq 0$. 
This implies that there exists a neighborhood of the desired formation’s equilibrium set such that $\lim_{t \rightarrow \infty}\widehat{\mathbf{L}}(\mathbf{x}) \mathbf{x} =0$ \cite{Krick2008},
given that $G$ is connected and (infinitesimally) rigid.
\end{proof}

\noindent
\textbf{Remark 1:}
Note that, as in standard consensus, the average state remains constant.
To see this, let $\hat{\mathbf{x}} = \frac{1}{N}\sum_{i = 1}^N \mathbf{x}_i$. 
Then,
\begin{align*}
    \dot{\hat{\mathbf{x}}} =  \frac{\alpha}{N}  \sum_{i = 1}^N \Big(\sum_{j \in \mathcal{N}_i}D_{ij}(\mathbf{x}_i - \mathbf{x}_j) + \sum_{j \in \mathcal{N}_i}D_{ij}(\mathbf{e}_i - \mathbf{e}_j) \Big) = 0,
\end{align*}
since $\sum_{i = 1}^N \sum_{j \in \mathcal{N}_i}D_{ij}\mathbf{x}_i = \sum_{i = 1}^N \sum_{j \in \mathcal{N}_i}D_{ij} \mathbf{x}_j$ and $\sum_{i = 1}^N \sum_{j \in \mathcal{N}_i}D_{ij}\mathbf{e}_i = \sum_{i = 1}^N \sum_{j \in \mathcal{N}_i}D_{ij} \mathbf{e}_j.$

\noindent
\textbf{Remark 2:}
The rigidity assumption in Theorem~\ref{thm:distributed_ctrl} is a sufficient condition~\cite{Krick2008}. 
In practice, the controller may converge to a zero-error configuration under a nonrigid graph, as illustrated in Section~\ref{sec:experiments}.

\subsection{Collision Avoidance} \label{sec:collision avoidance}

We employ distributed safety barrier certificates to guarantee collision avoidance among agents. 
The control input is obtained by solving a quadratic program (QP) that minimizes its deviation from the nominal input $\bar{\mathbf{u}}_i$, subject to safety constraints. 
Since the problem is a QP with linear constraints, it can be solved efficiently in real time \cite{Ames2019}.

Specifically, consider the control barrier function $h_{ij}(\mathbf{x}) = \delta_{ij}^2 - \delta_{\min}^2$, with the associated safe set $\mathcal{C}_{ij} = \left\{ \mathbf{x} \in \mathbb{R}^{\mathcal{D}N} \;\middle|\; h_{ij}(\mathbf{x}) \ge 0 \right\}.$
If the control input satisfies $\nabla_\mathbf{x} h_{ij}(\mathbf{x})^\top \mathbf{u} \ge -\gamma h_{ij}(\mathbf{x})^3$, for some $\gamma > 0$, then the set $\mathcal{C}_{ij}$ is forward invariant \cite{Wang2017}, thereby ensuring inter-agent safety.
The gradient of $h_{ij}$ is given by
\begin{equation}
\nabla_\mathbf{x} h_{ij}^q(\mathbf{x}) =
\begin{cases}
2(\mathbf{x}_i - \mathbf{x}_j), & q = i, \\
-2(\mathbf{x}_i - \mathbf{x}_j), & q = j, \\
0, & \text{otherwise}.
\end{cases}
\end{equation}
To obtain a distributed formulation, we distribute the constraint symmetrically between the two agents:
\begin{equation}
2(\mathbf{x}_i - \mathbf{x}_j)^\top \mathbf{u}_i \ge - \frac{1}{2} \gamma h_{ij}(\mathbf{x})^3,
\end{equation}
\begin{equation}
-2(\mathbf{x}_i - \mathbf{x}_j)^\top \mathbf{u}_j \ge - \frac{1}{2} \gamma  h_{ij}(\mathbf{x})^3.
\end{equation}
By aggregating these inequalities and enforcing the corresponding $|\mathcal{N}_i|$ constraints for every agent $i$, we ensure that all associated pairwise safe sets remain forward invariant.

The final control input for agent $i$ is obtained by
\begin{equation}
\mathbf{u}^*_i =
\arg\min_{\mathbf{u}_i \in \mathbb{R}^{\mathcal{D}}}
\left\| \mathbf{u}_i - \bar{\mathbf{u}}_i \right\|^2,
\end{equation}
subject to
\begin{subequations}\label{eq:collision_avoid_inequalities}
\begin{align}
    & -(\mathbf{x}_i - \mathbf{x}_j)^\top \mathbf{u}_i \le  \gamma \, h_{ij}(\mathbf{x})^3,
\quad \forall j \in \mathcal{N}_i, \\
    & -u_{\max}\mathbf{1} \le \mathbf{u}_i \le u_{\max}\mathbf{1},
\end{align}
\end{subequations}
where $u_{\max}$ is a maximum control bound.

The controller $\mathbf{u}_i^*$ deviates from the nominal controller $\bar{\mathbf{u}}_i$ only when collision avoidance constraints become active. 
The constraints in~\eqref{eq:collision_avoid_inequalities} are evaluated at the same time instants as the event-triggering condition in~\eqref{event_distr}.
Hence, $\mathbf{u}_i^*$ is a minimally invasive modification of $\bar{\mathbf{u}}_i$.
The following proposition establishes an upper bound on the deviation of the barrier-filtered controller from the nominal controller, which ensures that the total formation error will not increase.

\begin{proposition}\label{prop:coll_avoid}
Let $\mathbf{d}_i \triangleq \mathbf{u}_i^* - \bar{\mathbf{u}}_i$ and suppose that the event-triggering condition of Theorem~\ref{thm:distributed_ctrl} holds.
If at time $t$, the condition
\begin{align}
    \|\mathbf{d}_i(t)\| < \alpha A_i(\mathbf{x}(t), \mathbf{z}(t)),
\end{align}
holds for each agent $i$,
where $A_i$ is given by \eqref{eq:A_i}, then the total formation error will not increase.
\end{proposition}
\begin{proof}
Let $\mathbf{d}_i \triangleq \mathbf{u}_i^* - \bar{\mathbf{u}}_i$ and $\mathbf{d}$ be the stacked vector of $\mathbf{d}_i$.
Then, the closed-loop dynamics can be written as $\dot{\mathbf{x}} = \bar{\mathbf{u}} + \mathbf{d}$.
Using the Lyapunov function in \eqref{eq:LF}, we obtain
\begin{align}
    \dot{V} = \frac{\partial V} {\partial \mathbf{x}} \dot{\mathbf{x}} =  \frac{\partial V} {\partial \mathbf{x}} \bar{\mathbf{u}} + \frac{\partial V} {\partial \mathbf{x}} \mathbf{d}.
\end{align}

By Theorem~\ref{thm:distributed_ctrl}, the nominal controller satisfies 
$\frac{\partial V} {\partial \mathbf{x}} \bar{\mathbf{u}} \leq -\sum_i  A_{i} \| \mathbf{z}_i \| $.
Therefore,
\begin{align}
    \dot{V}
    \leq - \sum_{i} A_i \|\mathbf{z}_i\|
    + \sum_i \frac{1}{\alpha} \|\mathbf{z}_i\| \, \|\mathbf{d}_i\|.
\end{align}

If $\|\mathbf{d}_i\| < \alpha A_i$, it follows that $\dot{V} \leq 0$.
\end{proof}

\section{Experiments}\label{sec:experiments}

We conducted a variety of experiments and simulations to evaluate the proposed controller. 
We performed experiments using the Robotarium simulator and testbed \cite{Wilson2020_robotarium}, as well as a custom-built simulator for larger numbers of agents.

In the first part, we present experiments with different agent connectivity.
Next, we examine the impact of different design parameters.
Each experiment involves a team of six robots ($N = 6$), evenly spaced along a circle of radius 0.9 meters in $\Re^2$, as shown in Figure~\ref{fig:robotarium}(\subref{fig:robotarium_in}). 
The target formation is a 120-degree angle V-formation, where the robots are positioned at distances of 0.2, 0.6, and 1.0 meters from the vertex of the formation, as illustrated in Figure~\ref{fig:robotarium}(\subref{fig:robotarium_fin}). 

Note that Robotarium robots operate under nonholonomic constraints and are modeled using the unicycle dynamics.
To convert desired single-integrator commands to unicycle-compatible control inputs, we employ the near-identity diffeomorphism \cite{OlfatiSaber2002, Wilson2020_robotarium} using $\ell = 0.05$ meters.

Finally, we extend the study to three-dimensional simulations in $\Re^3$ with a significantly larger team of robots ($N = 200$). 
In this case, the robots are initially distributed over a Fibonacci sphere of radius 10, with neighbors defined by a 5-meter communication radius.
The objective here is not to achieve a specific formation but to ensure that if an agent has neighbors located on the opposite hemisphere, both agents move 20 meters apart. 
All other agents maintain their relative distances to preserve connectivity.

For the performance evaluation, for each agent $i$, we define 
$\tau_i^1$ as the total number of events generated by the triggering condition in \eqref{event_distr}, and 
$\tau_i^2$ as the total number of events due to collision avoidance. 
We then compute the average number of events across all agents  
$\tau^k = \frac{1}{N}\sum_{i=1}^{N}\tau_i^k,
\; k\in \{1,2\}.$

We also compute the average formation error 
$F(t) = \frac{1}{N} \sum_{i=1}^N  \frac{1}{|\mathcal{N}_i|} \sum_{j \in \mathcal{N}_i} |D_{ij}(t)|,$
which captures the deviation from the desired inter-agent distances.
To capture the formation error over the entire simulation, we also compute
$J_F = \int_{0}^{T} F(t)\,\mathrm{d} t,$
where $T$ denotes the final simulation time.

\subsection{Agent Communication Topology Variation}\label{sec:topology}

\begin{figure}[tb]
    \centering        
    {\includegraphics[width=0.75\linewidth, trim=90 50 55 30, clip]{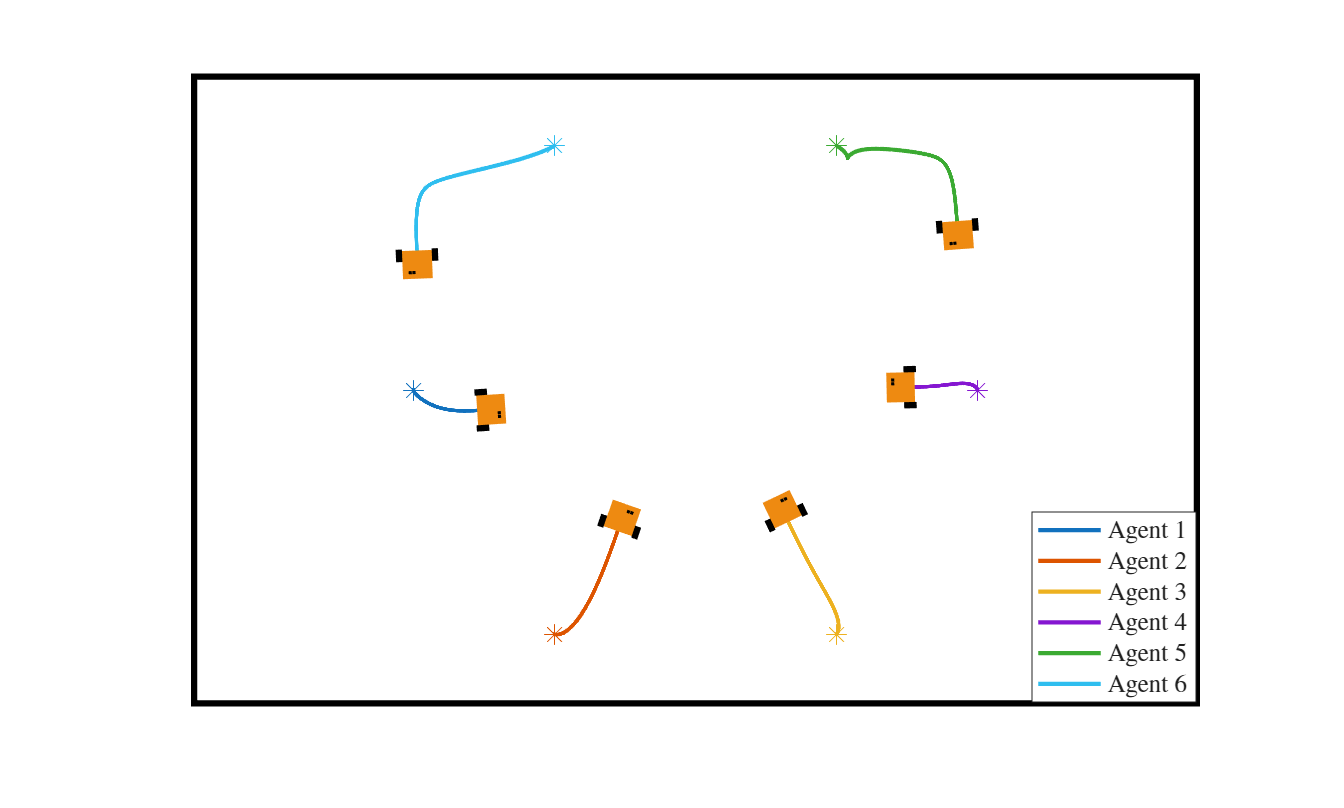}}
    \caption{Final positions and trajectories of agents at $t = 50s$ after applying the proposed controller with a cycle graph Laplacian. The symbol * indicates their initial positions.} 
    \label{fig:sim1_fin}
\end{figure}

\begin{figure}[tb]
     \centering
     \begin{subfigure}[b]{0.49\linewidth}
         \centering
         \includegraphics[width=\textwidth, trim=8 0 20 15, clip]{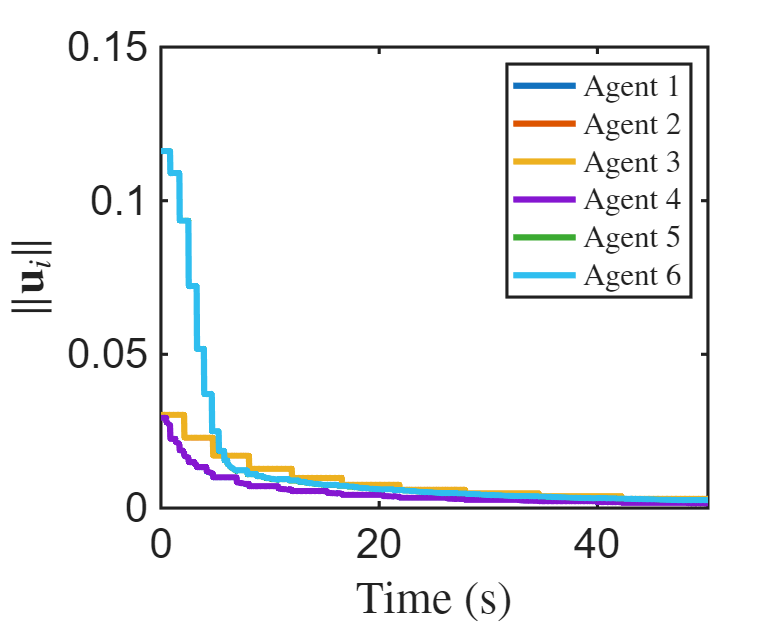}
         \caption{}
         \label{fig:control_cycle}
     \end{subfigure}
     \begin{subfigure}[b]{0.49\linewidth}
         \centering
         \includegraphics[width=\textwidth, trim=8 0 20 15, clip]{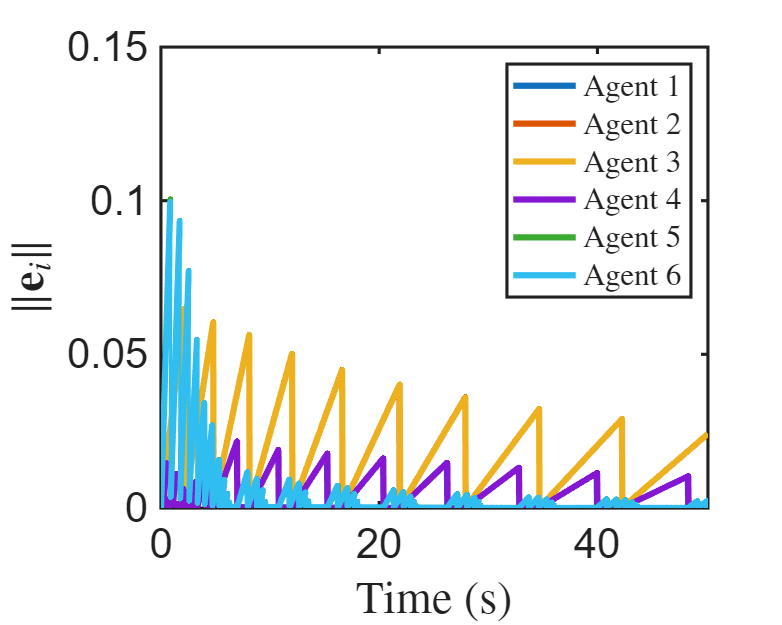}
         \caption{}
         \label{fig:error_cycle}
     \end{subfigure}

        \caption{Norms of the control inputs and measurement errors for six agents under a cycle-graph communication topology.}
        \label{fig:results_cycle}
\end{figure}

In this section, we apply the proposed controller in the Robotarium using a complete and a cycle graph.
The parameters were set to $\sigma_i = 0.9$ for all $i$, $\delta_{\min} = 0.35$, $\gamma = 100$, $T = 50s$ and $u_{\max}$ was set sufficiently large to showcase the effect of $\alpha$.
For the complete graph, the velocity gain was chosen as $\alpha = 0.01$, while for the cycle graph it was set to $\alpha = 0.05$.
A state-dependent gain can also be employed, but its study is beyond the scope of this work.

Figure~\ref{fig:sim1_fin} shows the agent trajectories obtained using the cycle-graph Laplacian, while Figure~\ref{fig:results_cycle} depicts the corresponding control-input norms and measurement errors for each agent \(i\). 
As expected, the control inputs are piecewise constant and converge to zero.
The measurement error exhibits oscillations whose frequency decreases over time.
Similar error and control plots were observed for the complete-graph topology. 
Finally, Figure~\ref{fig:robotarium} demonstrates the experimental deployment of the proposed controller on the physical Robotarium testbed using the complete Laplacian.

\begin{table}[!t]
\caption{Agent Communication Topology Variation.}
\label{tab:results1}
\centering
\small
\begin{tabular}{ |c|c|c|c|c| } 
    \hline
    {Metrics} &  {$\tau^1$} &  {$\tau^2$} & {$F(T)$} & {$J_F$} \\
    \hline
    {Complete $\mathbf{L}$ (ET)} & ${313}$ & ${4}$ & ${0.0727}$ & ${13.17}$ \\
    \hline
    {Complete $\mathbf{L}$ (PT, $T_s = 0.033$)} & \multicolumn{2}{c|}{1520} & 0.0819 & 14.20 \\
    \hline
    {Complete $\mathbf{L}$ (PT, $T_s = 0.33$)} & 152 & {7} & 0.0810 &  14.02 \\
    \hline
    {Cycle $\mathbf{L}$ (ET)} & $380$ & $0$ & 0.1036 & 11.15 \\
    \hline
    {Cycle $\mathbf{L}$ (PT, $T_s = 0.033$)} & \multicolumn{2}{c|}{1520} & 0.1093 & 11.62 \\
    \hline
    {Cycle $\mathbf{L}$ (PT, $T_s = 0.33$)} & {152} & 0 & 0.1087 & 11.53 \\
    \hline
\end{tabular}
\end{table}

Table~\ref{tab:results1} summarizes the performance of the event-triggered (ET) controller and two periodic-triggered (PT) controllers with different update periods. 
The ET controller reduces the number of control updates by approximately \(80\%\) relative to the PT controller with \(T_s=0.033\,\mathrm{s}\), while also achieving slightly better performance than both periodic baselines. 
The difference in \(J_F\) between the cycle- and complete-graph topologies is primarily due to the difference in \(\alpha\). 
Note that under the cycle Laplacian, the communication graph is not rigid.
However, the agents still converge to a configuration satisfying the desired inter-agent distances.

\begin{figure}[tb]
     \centering
     \begin{subfigure}[b]{0.45\linewidth}
         \centering
         \includegraphics[width=\textwidth, trim=30 0 70 50, clip]{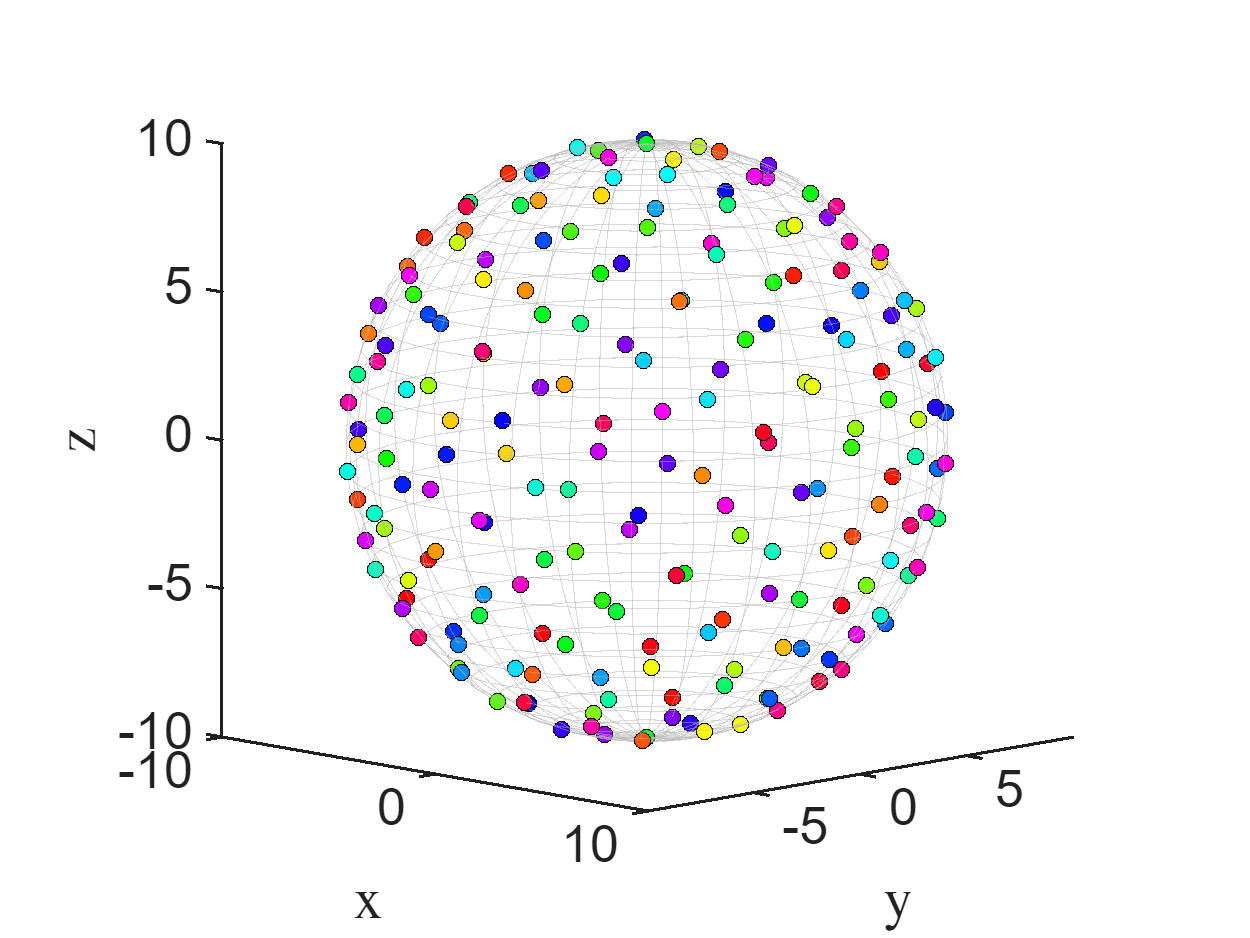}
         \caption{}
         \label{fig:large_pop_init}
     \end{subfigure}
     \begin{subfigure}[b]{0.51\linewidth}
         \centering
         \includegraphics[width=\textwidth, trim=10 0 40 40, clip]{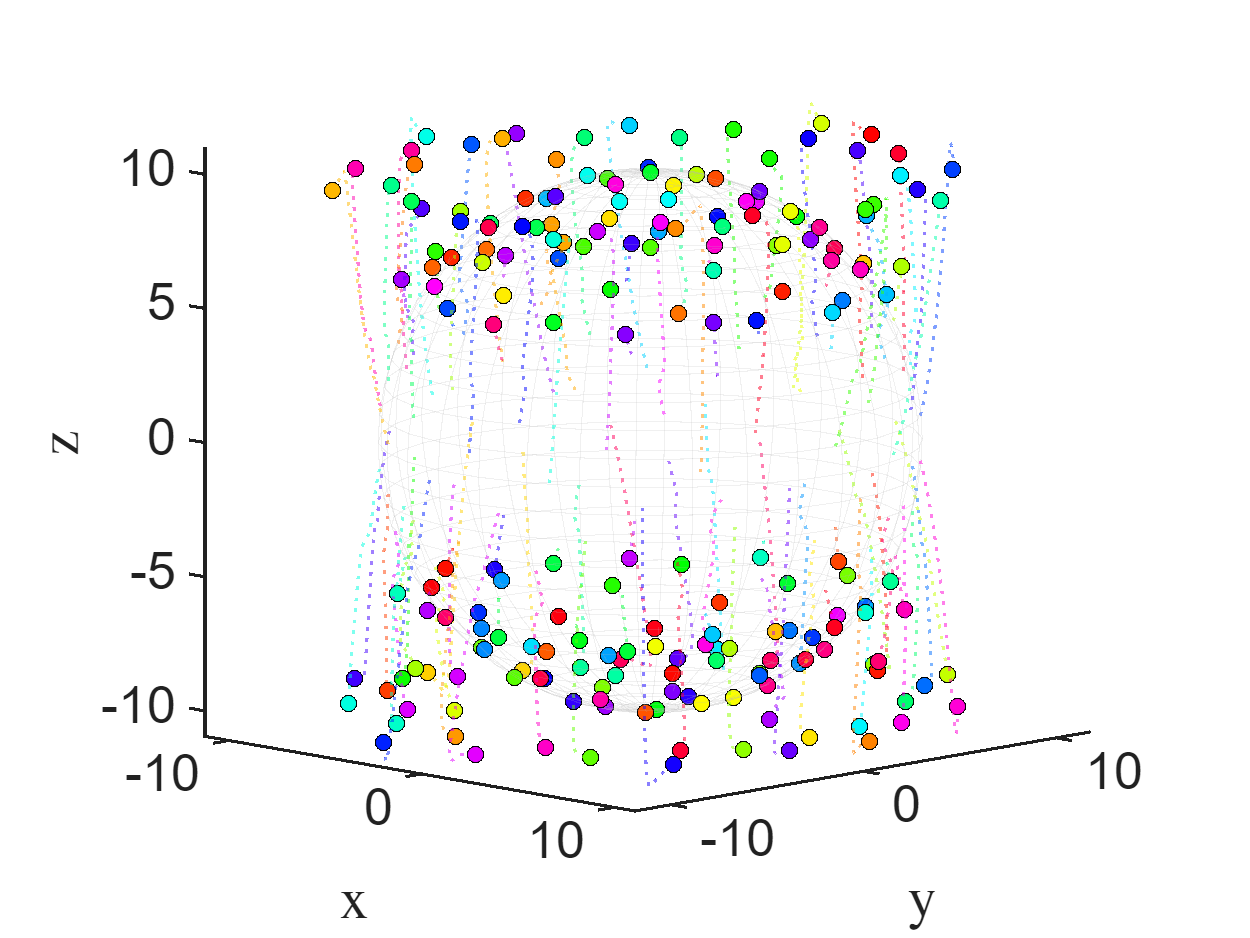}
         \caption{}
         \label{fig:large_pop_fin}
     \end{subfigure}
        \caption{Initial arrangement of 200 agents on a sphere (a) and their final positions (b), showing the increased separation between agents assigned to opposite hemispheres.} 
        \label{fig:large_pop}
\end{figure}
\subsection{Design Parameter Variation}

\begin{table}[!t]
\caption{Variation of $F(T)$ and $\tau = \tau^1 + \tau^2$ with respect to $\alpha$ and $\sigma$.}
\label{tab:results2}
\centering
\small
\begin{tabular}{|c|c|c|c|}
    \hline
    {$F(T)$/$\tau$} & {$\alpha = 0.01$} & {$\alpha = 0.05$} & {$\alpha = 0.1$} \\
    \hline
    {$\sigma = 0.01$}   & {0.2062/ 1345} & {0.0582/ 1962}  & \textit{0.0269/ 2234} \\
    \hline
    {$\sigma = 0.1$}   & {0.2052/ 706} & {0.0579/ 1094}  & \textit{0.0267/ 1271} \\
    \hline
    {$\sigma = 0.5$}   & {0.2010/ 593} & {0.0564/ 869}  & \textit{0.0260/ 974} \\
    \hline
    {$\sigma = 0.9$}   & {0.1972/ 573} & {0.0551/ 829}  & \textit{0.0253/ 923} \\
    \hline
    {$\sigma = 0.99$}   & {0.1966/ 537} & {0.0548/ 822}  & \textit{0.0252/ 918} \\
    \hline
\end{tabular}
\end{table}


In this section, we consider a cycle graph and investigate the effect of varying the parameters $\alpha$ and $\sigma_{i}$ in \eqref{event_distr}.
We set $\sigma_{i} = \sigma$ for all $ i$, $T = 100s$ and $\delta_{\min} = 0.15$.

\begin{figure}[!t]
    \centering        
    {\includegraphics[width=1.0\linewidth]{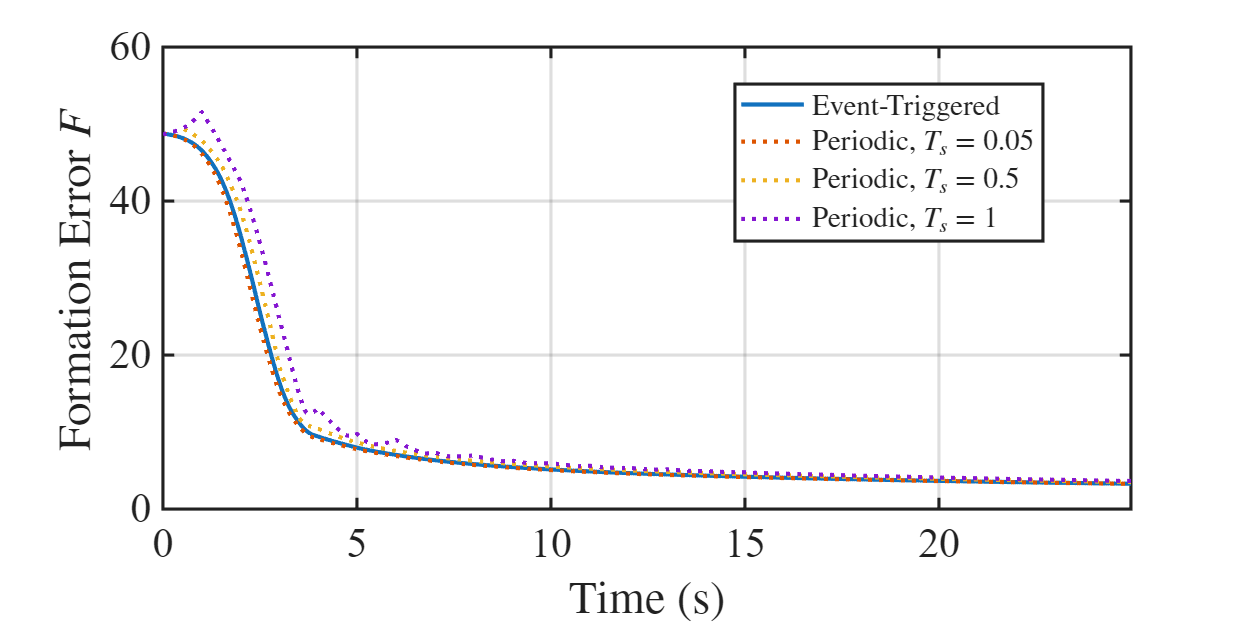}}
    \caption{Average formation error for a 200-agent simulation.} 
    \label{fig:results_large_pop}
\end{figure}

%
Table~\ref{tab:results2} summarizes the results.
Entries shown in italics correspond to cases where at least one of the agents' velocities exceeded the actuator limits.
Overall, increasing $\alpha$ improves formation performance, as indicated by the smaller values of $F(T)$, while increasing $\sigma$ reduces the number of triggered events, ($\tau^1 + \tau^2$).

\subsection{Scalability}

We evaluated the scalability of the proposed controller by running simulations with 200 agents in $\Re^3$.
The agents are initially placed on a sphere, as shown in Figure~\ref{fig:large_pop}(\subref{fig:large_pop_init}). 
The objective is for agents with neighbors in the opposite hemisphere to move 20 meters apart.
The design parameters were set as follows: $\sigma_i = 0.5$ for all $i$, $\alpha = 0.00025$, $\delta_{\min} = 1.6$, $\gamma = 100$, and $u_{\max} = 10$.

Figure~\ref{fig:large_pop}(\subref{fig:large_pop_fin}) shows the final agent positions and trajectories, indicating that the desired objective is achieved.
Figure~\ref{fig:results_large_pop} further supports this observation by showing that the formation error decreases consistently under the ET controller.
Figure~\ref{fig:results_large_pop} also compares the ET controller with three PT baselines having sampling periods $T_s = 0.05$, $T_s = 0.5$, and $T_s = 1$, corresponding to $\tau^1 + \tau^2 =  500$, $54$, and $34$ update counts, respectively. 
The average formation error obtained with the ET controller is nearly identical to that of the PT controller with $T_s = 0.05$, while both outperform the PT baselines with larger sampling periods. 
However, the ET controller requires substantially fewer updates, with average triggering counts of $\tau^1 = 354$ and $\tau^2 = 2$, compared with the $500$ updates required by the PT controller with $T_s = 0.05$. 

\section{Discussion on Zeno Behavior}

A well-known challenge for implementing an event-triggered controller is the possibility of Zeno behavior. 
In our simulations, this phenomenon was not observed (see Figure~\ref{fig:results_cycle}(\subref{fig:error_cycle})). 
Nevertheless, providing a formal proof of Zeno exclusion for the proposed triggering rule is nontrivial due to the dynamic and coupled structure of the triggering law.

That said, methods have been developed in the literature that can be leveraged to address this issue. 
For instance, periodic event-triggered laws~\cite{NOWZARI2016} introduce a sampling period to evaluate the triggering condition, which naturally precludes Zeno behavior and is also well aligned with the constraints of real-time digital implementations.
Incorporating such a modification into the proposed triggering law is a promising direction for future work.

\section{Conclusions}

We have proposed a distributed, event-triggered, distance-based formation controller for multi-agent systems with limited resources.
We established the stability of the controller and used barrier certificates for collision avoidance among the agents.
Experimental results show that the proposed controller substantially reduces the number of control updates while maintaining formation accuracy.
For future work, we plan to extend the framework beyond single-integrator dynamics by incorporating learning-based techniques to capture unknown system dynamics.





\bibliographystyle{IEEEtran}
\bibliography{IEEEabrv,refs}

\end{document}